\documentclass{article}

\usepackage{microtype}
\usepackage{graphicx}
\usepackage{caption}
\usepackage{subcaption}
\usepackage{booktabs} 
\usepackage{amsfonts}

\PassOptionsToPackage{hyphens}{url}\usepackage{hyperref}

\usepackage[accepted]{icml2019}

\usepackage{todonotes}

\icmltitlerunning{How do MD-RNNs predict the future?}

\begin{document}

\twocolumn[
\icmltitle{How do Mixture Density RNNs Predict the Future?}



\icmlsetsymbol{equal}{*}

\begin{icmlauthorlist}
  \icmlcorrespondingauthor{Kai Olav Ellefsen}{uio}
  \icmlauthor{Kai Olav Ellefsen}{uio}
  \icmlauthor{Charles Patrick Martin}{uio,ritmo}
  \icmlauthor{Jim Torresen}{uio,ritmo}
\end{icmlauthorlist}

\icmlaffiliation{uio}{Department of Informatics, University of Oslo, Norway}
\icmlaffiliation{ritmo}{RITMO, University of Oslo, Norway}
\icmlkeywords{Prediction, MD-RNNs}

\vskip 0.3in
]



\printAffiliationsAndNotice{}  


\begin{abstract}
  Gaining a better understanding of how and what machine learning systems learn 
  is important to increase confidence in their decisions and catalyze further
  research. In this paper, we analyze the predictions made by a specific type of
  recurrent neural network, mixture density RNNs (MD-RNNs).
  These
  networks learn to model predictions as a combination of multiple
  Gaussian distributions, making them particularly interesting for
  problems where a sequence of inputs may lead to several distinct
  future possibilities.
  An example 
  is learning internal models of an environment, where different events may or 
  may not occur, but where the average over different events is not 
  meaningful. By analyzing the predictions made by trained MD-RNNs, we find 
  that their different Gaussian components have two complementary roles: 1) 
  Separately modeling different stochastic events and 2) Separately modeling 
  scenarios governed by different rules. These findings increase our 
  understanding of what is learned by predictive MD-RNNs, and open up new 
  research directions for further understanding how we can benefit from their 
  self-organizing model decomposition.
\end{abstract}



\section{Introduction}
\label{S:1}

Deep learning has greatly increased the ability for computers to
perform complex tasks from a wide range of domains, including image
recognition, language modeling, game playing and predicting the
future~\cite{Mnih2015, LeCun2015, Wichers2018}. However, we have an
incomplete understanding of exactly how the deep learning models 
learn to perform these tasks. Gaining a better understanding
of these models is considered to be one of the most important current
challenges in artificial intelligence~\cite{samek2017explainable,
  garcia2018task}. There are many reasons why a better understanding
is important, ranging from increasing the ability to trust machine
learning systems, to the benefits such understanding would have for
continued research and development of algorithms. There are therefore
many recent studies aiming to make deep learning models more
understandable and explainable, for both convolutional neural
networks (CNNs)~\cite{yosinski2015understanding}, recurrent neural
networks (RNNs)~\cite{karpathy2015visualizing} and other
architectures~\cite{smilkov2017direct}.

\begin{figure*}
  \begin{center}
    \includegraphics[width=\textwidth]{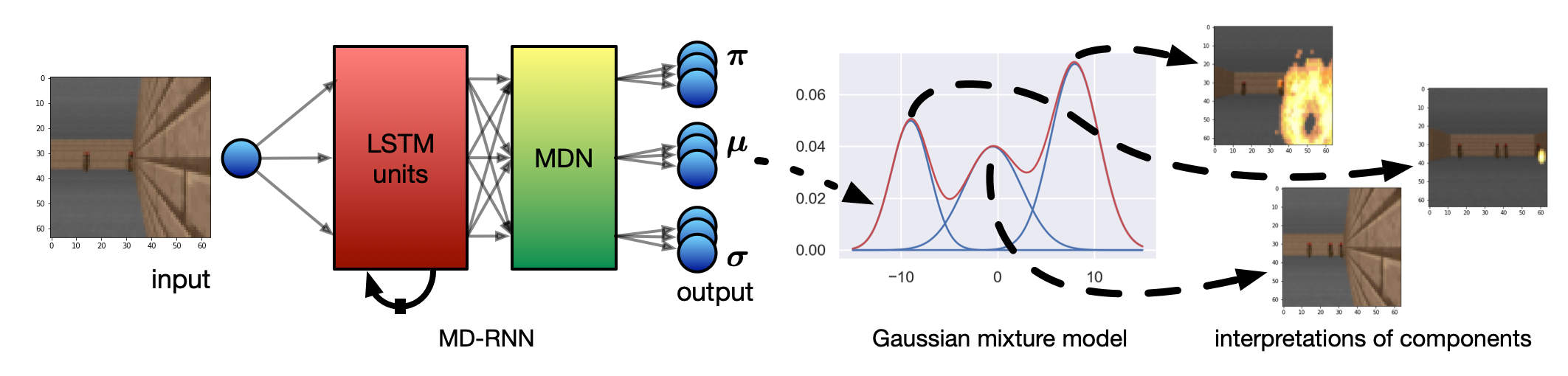}
    \caption{Left: MD-RNNs model data with probability distributions 
    composed of several components, parametrized by $\pi$, $\mu$ and $\sigma$. 
    Right: We investigate the roles of individual components to gain a better 
    understanding of how MD-RNNs make predictions. One possibility 
      (illustrated here) is that different mixture components represent 
      situations governed by different rules.}
    \label{fig:mdn_motivations}
  \end{center}
\end{figure*}

In this paper, we aim to gain a better understanding of one specific
neural network architecture, mixture density RNNs (MD-RNNs). MD-RNNs are 
recurrent neural
networks combined with a mixture density network~\cite{Bishop1994,
  bishop1995neural}, such that that the output parametrizes a mixture
of Gaussians distribution (Figure~\ref{fig:mdn_motivations}). MD-RNNs are 
particularly interesting for tasks
involving creative prediction, since the recurrent part allows the
modeling and forecasting of sequences, and the Gaussian mixture part
allows predictions to be creative, modeling different types of
scenarios in a single neural network~\cite{Ha2013}.



MD-RNNs recently gained 
significant attention~\cite{Pearson, Yao} in Ha and
Schmidhuber's paper on ``World Models''~\cite{NIPS2018_7512}, which
demonstrated that MD-RNNs, can learn to predict the future from a
large number of observations of a simulated world. The internal models learned 
in Ha and 
Schmidhuber's work represent
the agent's world so well that the authors were able to 1) train an
agent \emph{inside its own internal model} (or, said differently,
inside its own ``dream'') and 2) to be the first agent
to solve the Car Racing environment in OpenAI gym. Despite the
impressive results and the wide attention this work has gotten, we do
not have a good understanding of how predictive MD-RNNs model the
world.

MD-RNNs make predictions by sampling from a probability distribution with 
multiple different sub-distributions. We investigate two hypotheses about the 
role of these sub-distributions (mixture components) when MD-RNNs predict the 
future. The hypotheses are:
1) Different mixture components model different stochastic events and 2)
Different mixture components model different situations with different
``rules'' (that is, different internal models -- 
Figure~\ref{fig:mdn_motivations}). We train world models in a Doom
game environment, similarly to~\cite{NIPS2018_7512}, and let them hallucinate 
imagined scenarios.
From these scenarios, we extract events and situations according to
our two hypotheses, and then measure to which degree different events
are produced by different components of the mixture model.


The main contributions of this paper are 1) A framework for
automatically measuring the tendency for different components of a
Gaussian mixture model to generate particular types of prediction, and
2) New insights into the roles of the Gaussian components of trained 
MD-RNNs. In particular, we find evidence for both our hypotheses, including a 
very clear demonstration of different  mixture components self-organizing to 
serve as internal models for scenarios with different rules.

\section{Background}

\subsection{MD-RNNs}

Generative machine learning models for content such as text, images or sound 
typically model the generated content with a probability 
distribution~\cite{goodfellow2016deep_20}. Mixture density networks (MDNs) 
are neural networks that represent mixture 
density models~\cite{mclachlan1988mixture}, that is, probability distributions 
which are composed of several sub-distributions (several Gaussian distributions 
in the models applied here -- see Figure~\ref{fig:mdn_motivations}). MDNs can 
in principle represent any conditional probability distribution, and are useful 
when the modeled phenomenon is not well represented by a simpler distribution. 
An example, which we study here, 
is learning internal models of an environment, where different events may or 
may not occur, but where the average over different events is not 
meaningful. In this case, the multiple sub-distributions of a mixture density 
model can help model the fact that the world has multiple possible states which 
should not be mixed together or averaged.

In practice, a mixture density network (MDN) operates by transforming the 
outputs of a neural
network to form the parameters of a mixture distribution~\cite{Bishop1994},
generally with Gaussian models for each mixture component. These
parameters are the centres ($\mu$) and scales ($\sigma$) for each
Gaussian component, as well as a weight ($\pi$) for each component
(see Figure \ref{fig:mdn_motivations}). The MDN usually uses an
exponential activation function to transform the scale parameters to
be positive and non-zero. For training, the probability density
function of the mixture model is used to generate the negative log
likelihood for the loss function. This involves constructing probability 
density functions (PDFs) for
each Gaussian component and categorical distribution from the mixture
weights (see Appendix Section 1.4 for details). One advantage of an MDN is that 
various component
distributions can be used so long as the PDF is tractable, for
instance, 
1D~\cite{Bishop1994} or 2D~\cite{Graves:2013aa} Gaussian
distributions, or, as in our case, a multivariate Gaussian with a
diagonal covariance matrix.

For inference, results are sampled from the mixture distribution.
First, the $\pi$s are used to form a categorical distribution by
applying the softmax function. A sample is drawn from this
distribution to determine which Gaussian component will provide the
output. The index $i$ of the sampled $\pi$ is used to select a Gaussian
distribution, $\mathcal{N}(\mu_i, \sigma_i^2)$, from which a sample is
drawn to provide the outcome. In some cases, it is advantageous to
adjust the diversity of sampling (for instance, to favour unlikely
predictions), in which case the temperature of the categorical
distribution  can be adjusted in the
typical way, and the covariance matrices of the Gaussian components
may be scaled. We refer the these operations as adjusting $\pi$- or
$\sigma$-temperature respectively.

An MDN can be applied to the outputs of an RNN,
forming an MD-RNN. This approach has been applied to model 2D pen
data, such as for handwriting~\cite{Graves:2013aa} and
sketches~\cite{Ha2013} as well as musical
performance~\cite{Martin2018}. Other applications include parametric
speech synthesis~\cite{Wang2017}, and identifying salient locations in
video data~\cite{Bazzani:RMDNattention}. Ha and Schmidhuber applied an
MD-RNN model to model the future state of a video game screen image
and assist an RL agent~\cite{NIPS2018_7512}. In the present research,
we delve into this application to understand what such a model learns
about the virtual worlds and how this information is represented.

\subsection{Predicting the future with deep neural networks}


Progress in deep learning has recently made it possible to learn to
predict future frames of video from observing sequences of
video frames~\cite{2016arXiv160507157F, Mathieu2016}. However, most
approaches for predicting future visual input from pixels
have typically only had the ability to predict a few frames into the
future before predicted images get blurry or static. Recently,
techniques have been developed that attempt to mitigate this
limitation by first encoding frames into a compact, high-level
representation, then predicting how this compact representation
develops over time. Finally, decoding the predicted compact
representation produces a predicted future image. Compared to
predictions made directly in pixel-space, such high-level predictions
degrade less quickly, demonstrating good prediction performance many
seconds into the future~\cite{2017arXiv170405831V, Wichers2018,
  NIPS2018_7512}.

In Ha and Schmidhuber's recurrent world model~\cite{NIPS2018_7512}, the 
predictive model consists of two components: 1) A visual component
(V), which learns an encoding/decoding between a visual scene and a
compact representation and 2) A memory component (M), which learns how
the compact representation develops over time (Figure~\ref{fig:world_model}). 
The first component is
learned by a variational autoencoder (VAE~\cite{kingma2013auto}), by
presenting it with a large collection of pictures from the visual
scene. The second is learned by an MD-RNN. This
world model was demonstrated to be able to predict many frames into
the future, and in fact to ``dream'' whole episodes of agent
experience. It is, however, not
clear what role the different mixture components in the MD-RNN play
in predicting the future.

\subsection{Architectures for multiple internal models}

One of our hypotheses suggests that the different Gaussian components
learn to model different situations with different ``rules'', that is,
situations where predictions need to be so different that they are
best modeled separately. Humans show a remarkable ability to learn
internal models (mental simulations) of a wide range of different
situations, objects and people, without a high degree of conflict or
interference between them.

One theory suggests that this ability is facilitated by the modular
organization of our central nervous system. Neural modularity may be a
key to allow multiple internal models to coexist, enabling the
selection of the appropriate actions for the current
context~\cite{Ghahramani1997,Wolpert2003}. Computational models built around 
this
idea have indeed demonstrated the ability to learn and maintain
multiple internal models, and select the appropriate model for a given
context~\cite{Wolpert1998b, Haruno2001,Demiris2006}. These models work by
dividing learning experiences into multiple modules, where different
modules \emph{compete} to represent different situations. After a
number of learning episodes, this causes different modules to
specialize at representing different internal models, allowing the
system to model situations with different rules, with minimal
interference. Our hypothesis suggests that the Gaussian components of
the MDN self-organize to perform a similar task, allowing scenarios
with different rules to be modeled with little interference.


\section{Methods}

\subsection{World Model MD-RNN}

\begin{figure}
  \vskip 0.2in
  \begin{center}
    \centerline{\includegraphics[width=0.6\columnwidth]{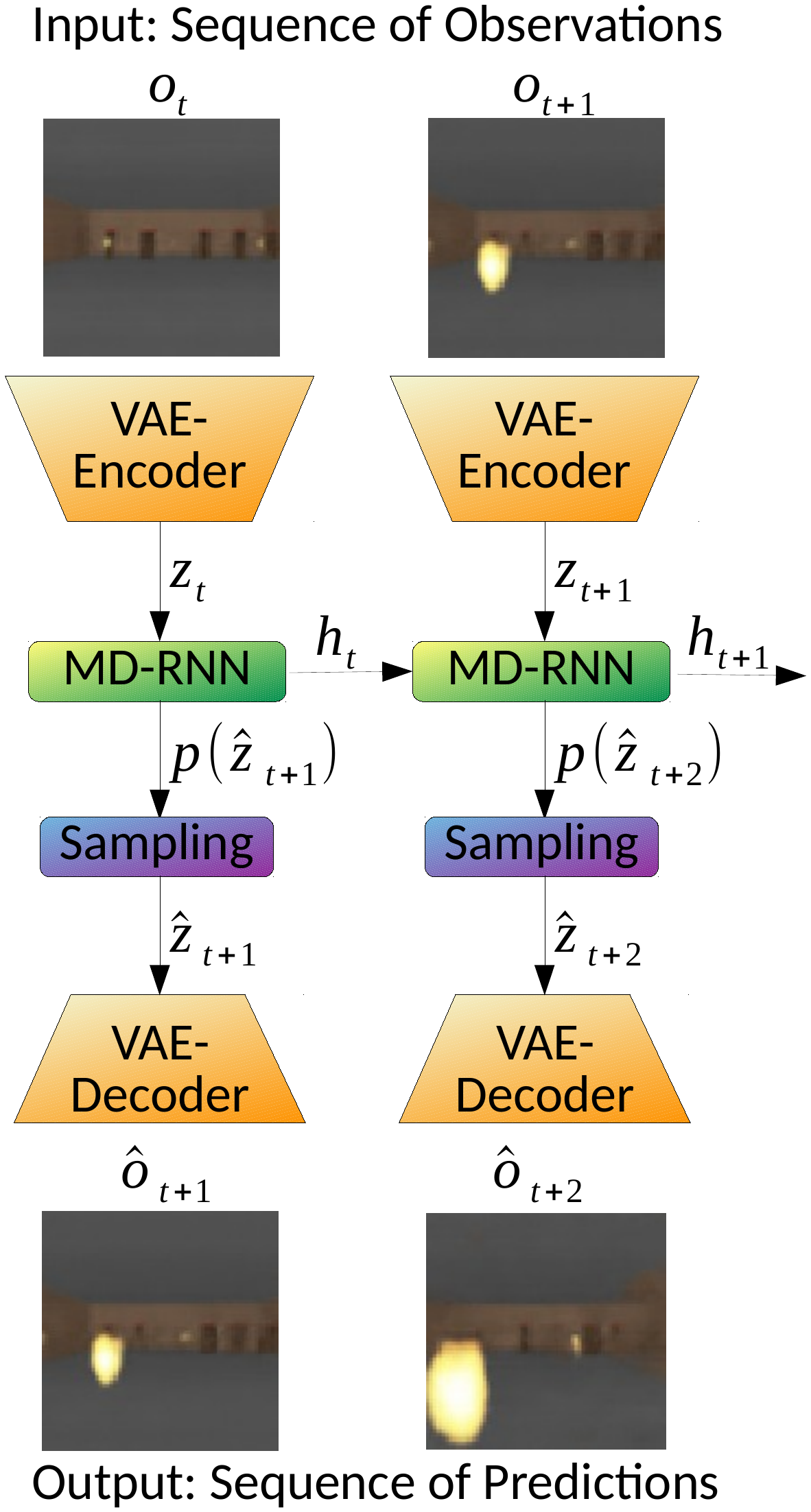}}
    \caption{World model predicting future frames by combining a 
      variational 
      autoencoder and an MD-RNN. We follow the architecture 
      suggested 
      in~\cite{NIPS2018_7512}.}
    \label{fig:world_model}
  \end{center}
  \vskip -0.2in
\end{figure}

Ha and Schmidhuber's world model~\cite{NIPS2018_7512} combines an
MD-RNN and a VAE to predict future states of a video game screen
(Figure~\ref{fig:world_model}). By training the VAE to 
compress representations of visual scenes, the MD-RNN has a more
manageable job of predicting how scenes unfold in the future.

Training the model happens in two steps: First, the VAE is trained on
examples of images from the environment in which we wish to learn to
make predictions. The VAE compresses each image (64x64 pixels, with 3
color channels in our setup) into a latent vector, $z$ (64
floating-point numbers in our case). It then attempts to reconstruct
the same image from the latent vector. The VAE is trained to both reconstruct 
images as well as possible, and to keep the representations of similar inputs 
close together in latent space (details are found in Appendix Section 1.3). 
This allows 
small changes to the latent vector to give meaningful changes in the compressed 
images.


After the VAE has learned to compress images of the world
into latent vectors, the MD-RNN can be trained on sequences
of latent vectors. We
follow~\cite{NIPS2018_7512} in applying a single-layer 
LSTM~\cite{Hochreiter1997}, trained
by seeing examples of sequences of images as input, and the same
sequence, shifted by one time-step, as outputs. Thereby, the LSTM
learns to predict the next latent vector from a sequence of previous
observations. More details on the World Model MD-RNN are found in the Appendix 
Section 1.4.


\subsubsection{Data collection and training}

The data collection and training process 
follows~\cite{NIPS2018_7512}, except we do not train a controller, since we are 
here \emph{analyzing} predictions, and not using them for agent control. The 
process can be summarized in the following 
steps (more details in Appendix 
Section 1\footnote{Experiment code is available at: 
\url{http://doi.org/10.5281/zenodo.2539145}}.

\begin{enumerate}
  \item Simulate 2,000 episodes with a random policy. Store all actions taken 
  and frames observed.
  \item Train a VAE to encode each frame into a length 64 latent 
  vector $z$, and to decode $z$ back to the same image.
  \item Generate latent vectors $z$ for each frame from the simulated episodes. 
  Further training can now be done without the actual images.
  \item Train an MD-RNN to model $P(z_{t+1} | a_t, z_t, h_t)$, that is, the 
  probability distribution for next latent vector, given the current latent 
  vector and action, as well as the RNN's hidden state. 
\end{enumerate}

\subsection{Training Scenario}

We follow~\cite{NIPS2018_7512} in training the predictive MDN-RNNs to model the 
VizDoom~\cite{Kempka2017} Take Cover 
scenario\footnote{\url{https://gym.openai.com/envs/DoomTakeCover-v0/}}. This 
scenario takes place in a rectangular room, where a player is facing monsters 
on an opposite wall. Monsters will fire exploding fireballs at the player, and 
the player attempts to survive as long as possible by moving left and right, 
dodging the incoming projectiles. Agents receive 3D images of the 
scene ahead of them as input, and make only one decision at each timestep: Move to the 
left, move to the right or stay in the same place.

This scenario serves as a useful test for our hypotheses, since it has both 
stochastic events (e.g., monsters may or may not launch fireballs) and different 
situations governed by different rules (an exploding fireball behaves very 
differently than an incoming fireball mid-air). 

\subsection{Measuring how MD-RNNs make predictions}

\begin{figure*}
  \vskip 0.2in
  \begin{center}
    \centerline{\includegraphics[width=\textwidth]{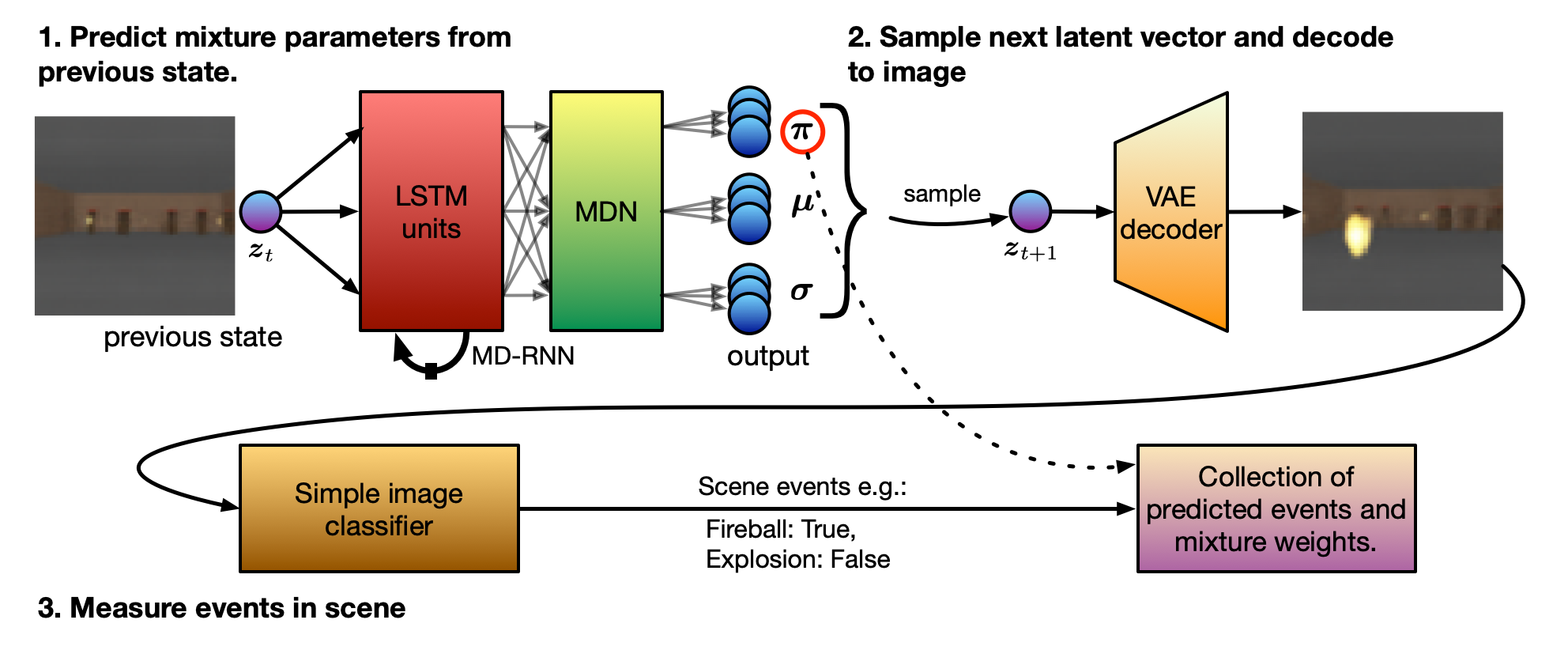}}
    \caption{Our proposed framework for analyzing how MD-RNNs make 
    predictions}
    \label{fig:framework_cropped}
  \end{center}
  \vskip -0.2in
\end{figure*}


After training MD-RNNs, we analyze the predictions they make when
``dreaming'' about the future. We insert an initial latent
vector (representing the real initial state from the game) into the MD-RNN, and 
then repeat the steps below for as long
as we want to predict (Figure~\ref{fig:framework_cropped} illustrates the 
steps, 
following the same numbering as the list):

\begin{enumerate}
  \item Produce a probability distribution over the next latent vector, 
  $P(\hat{z}_{t+1} | a_t, \hat{z}_t, h_t)$ parametrized by the MDN-parameters, 
  $\pi, 
  \mu$ and $\sigma$. Store $\pi$ (the vector indicating the weight of each 
  Gaussian component).
  \item Sample a latent vector $\hat{z}_{t+1}$ from the probability 
  distribution, and 
  decode it into a predicted frame with the VAE.
  \item Analyze the predicted frame to measure which events are depicted (see 
  below). Store the list of events for the current frame together with $\pi$. 
  Together, these can tell us whether different mixture components generate 
  different events.
  \item Repeat the process, starting from point 1, with the sampled 
  $\hat{z}_{t+1}$ as the RNN input, to predict the next latent 
  vector $P(\hat{z}_{t+2} | a_{t+1}, \hat{z}_{t+1}, h_{t+1})$
\end{enumerate}

Every round through this process generates a new predicted latent vector, which 
is next used as input to predict the latent vector following it. Storing latent 
vectors and MD-RNN parameters allows us to subsequently analyze the way the 
MD-RNN has learned to represent the world and make predictions about it.

\subsubsection{Measuring events in predicted frames}

Our hypotheses suggest that different mixture components represent
either different stochastic events, or different situations where
different rules apply. In the world we make predictions about, we
identify two stochastic events: 1) Monsters may appear, and 2) they
may launch a fireball towards the player. Note that monsters never
disappear in the modeled world, and fireballs disappearing is not a
stochastic event, since once a fireball has been fired, it will
deterministically disappear after reaching the other end of the room.

\begin{figure}
  \centering
  \begin{subfigure}[b]{0.24\columnwidth}
    \includegraphics[width=\textwidth, trim={1cm 1cm 0 
      0},clip]{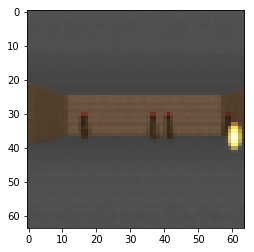}
    \label{fig:propagate1}
  \end{subfigure}
  \hfill
  \begin{subfigure}[b]{0.24\columnwidth}
    \includegraphics[width=\textwidth, trim={1cm 1cm 0 
      0},clip]{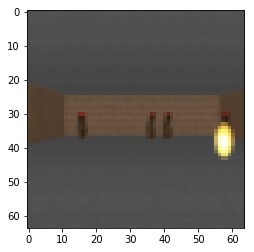}
    \label{fig:propagate2}
  \end{subfigure}
  \hfill
  \begin{subfigure}[b]{0.24\columnwidth}
    \includegraphics[width=\textwidth, trim={1cm 1cm 0 
      0},clip]{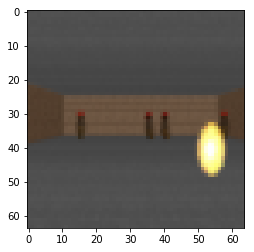}
    \label{fig:propagate3}
  \end{subfigure}
  \hfill
  \begin{subfigure}[b]{0.24\columnwidth}
    \includegraphics[width=\textwidth, trim={1cm 1cm 0 
      0},clip]{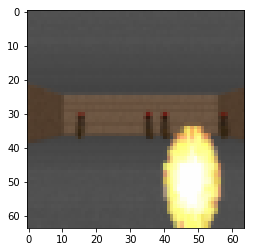}
    \label{fig:propagate4}
  \end{subfigure}
  
  \begin{subfigure}[b]{0.24\columnwidth}
    \includegraphics[width=\textwidth, trim={1cm 1cm 0 
      0},clip]{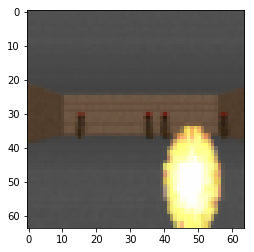}
    \label{fig:propagate1}
  \end{subfigure}
  \hfill
  \begin{subfigure}[b]{0.24\columnwidth}
    \includegraphics[width=\textwidth, trim={1cm 1cm 0 
      0},clip]{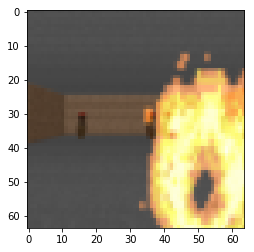}
    \label{fig:propagate2}
  \end{subfigure}
  \hfill
  \begin{subfigure}[b]{0.24\columnwidth}
    \includegraphics[width=\textwidth, trim={1cm 1cm 0 
      0},clip]{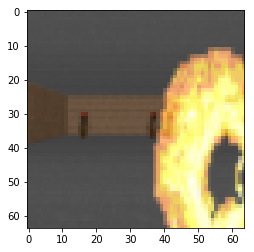}
    \label{fig:propagate3}
  \end{subfigure}
  \hfill
  \begin{subfigure}[b]{0.24\columnwidth}
    \includegraphics[width=\textwidth, trim={1cm 1cm 0 
      0},clip]{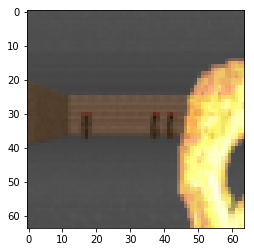}
    \label{fig:propagate4}
  \end{subfigure}
  \caption{Top: A monster launching a fireball at the player. Bottom: An 
    explosion unfolding in front of the player. The two situations are 
    governed 
    by very different rules. Images down-sampled to the same resolution 
    (64x64) 
    used during training.}
  \label{fig:propagation_vs_explosion}
\end{figure}

For our second hypothesis, we identify three situations where the rules for how 
frames evolve
in a time sequence are very different: 1) The normal situation (player
is facing monsters, who sometimes launch fireballs), 2) an explosion
takes place in front of the player and 3) the player is next to a
wall. Situation 2 and 3 are so different from the normal situation
that internal models of the three different situations could benefit
from some separation. Explosions cover a large portion of the screen,
and unfold according to a specific sequence, which has little to do
with the way a normal scene unfolds (see 
Figure~\ref{fig:propagation_vs_explosion}). Walls next to the 
agent result in
unique dynamics, since they require a large portion of the screen to
move sideways (in the opposite direction) as the player moves.

Since we are dealing with a quite simple and limited world, we can
measure events from frames with straightforward image processing
methods from the Python package
scikit-image\footnote{\url{https://scikit-image.org/}}. The methods we apply to 
measure the presence of monsters, fireballs, walls
and explosions are documented in Appendix Section 2, and also made available
online\footnote{\url{http://doi.org/10.5281/zenodo.2539145}}.

\section{Results}


As previously discussed, we have two main hypotheses about the roles
of different mixture components in the MD-RNN: 1) Different
components learn to model different possible futures, allowing them to
creatively sample what will happen next, and 2) Different components
learn to form different \emph{internal models} of the environment,
that is, they specialize to model situations governed by a specific
set of rules. Below, we analyze MD-RNN predictions along with the
weights of mixture components to shed light on these hypotheses.

\subsection{Common parameters}

In our main experiments, we test 5 independently trained MD-RNNs, all
with the same architecture (see Appendix Section 1), to reduce the chance
that results are specific to one trained model. In practice, we found
results to be very similar when training the same model multiple times
with shuffled data. For all five, we generate multiple ``dreams'' by
predicting future latent vectors, and feeding each prediction in as
the input-vector to the RNN for the next time-step, along with a
randomly sampled action. This allows us to dream up long prediction
sequences which, although not always realistic, illuminate how mixture
components relate to predicted events. In our main experiments, we
generate 10 dreams for each of our 5 models, each dream 1000
time-steps long.  Tests of statistical significance apply the Mann-Whitney U 
test.


\subsection{Analyzing frames produced in prediction sequences}

\subsubsection{Stochastic events}

Our first hypothesis suggests that different mixture components
represent different stochastic events, allowing creative predictions
about the future by sampling from different Gaussian components. We test this
hypothesis by dreaming up many different futures as described above,
and measuring 1) different \emph{stochastic events} in the dreams and
2) the weight assigned to each component in the mixture model. As
mentioned above, there are two different stochastic events in this
scenario: fireballs appearing and monsters appearing.

\begin{figure}[ht]
  \vskip 0.2in
  \begin{center}
    \centerline{\includegraphics[width=\columnwidth]{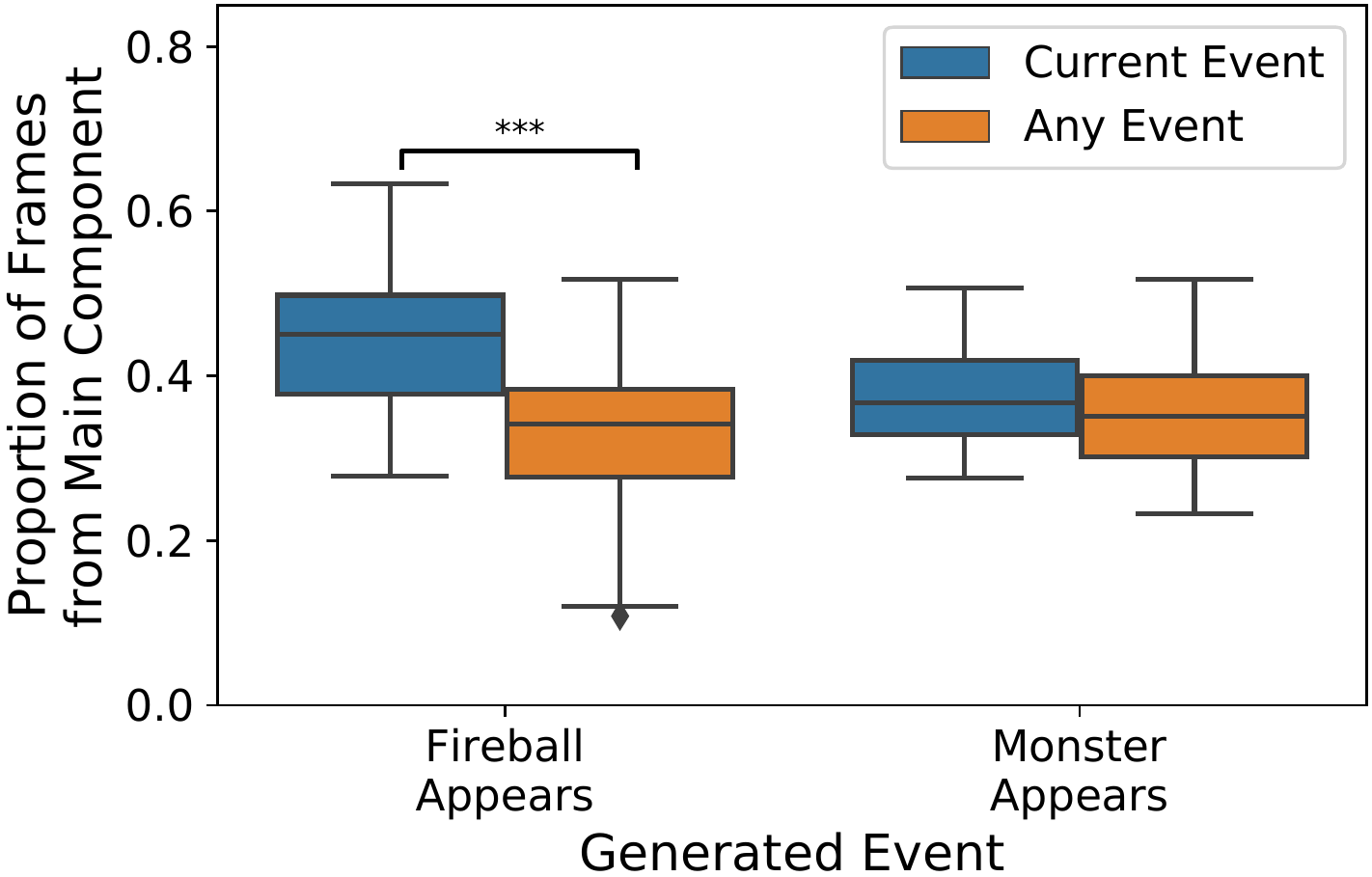}}
    \caption{The tendency for different stochastic events to be produced by one 
    specific Gaussian component (blue) vs the tendency for that component to be 
    responsible for events overall (orange).~$***$~indicates significant 
    differences with $p<0.001$.}
    \label{fig:event_proportions}
  \end{center}
\end{figure}

To confidently say that a specific mixture component is particularly
responsible for producing one event, we need to measure whether that
component has produced the event more frequently than one would expect
if events were evenly distributed among components. For instance, if
we find that one component is responsible for 80\% of the fireball
appearances, but that component is also responsible for 80\% of all
generated frames, then we do not have any clear evidence. We therefore
measure the relationship between components and events as follows:

\begin{enumerate}
  \item Produce 10 different dreams with each of the 5 trained 
  MD-RNNs, resulting in a total of 50 dreams.
  \item For each time-step of a dream, measure a) the presence of the events 
  described above, and b) which component is currently the most active (the one 
  with the highest $\pi$-value output by the MD-RNN).
  \item Within one dream, the component that produced an event most frequently, is denoted as the ``main component'' for that event. This is the Gaussian that is most likely responsible for generating the given event.
  \item Measuring the proportion of the event produced by the ``main component'' versus the other components across all N dreams yields the leftmost boxes in the pairs in Figure~\ref{fig:event_proportions}.
  \item To be sure the ``main component'' is specifically responsible for the 
  specific event/situation, we also measure the proportion of \emph{all} frames 
  produced by that component. This yields the rightmost boxes.
  \item A significantly higher value in the leftmost than the rightmost box thus indicates that one component is producing the relevant event/situation more frequently than one would expect by looking at the proportion of all events generated by that component.
\end{enumerate}

As we can see in Figure~\ref{fig:event_proportions}, there is a strong
tendency for fireball appearances to be produced more by one specific
component. There is no similar tendency for monster appearances.

\subsubsection{Different internal models}

Our second hypothesis is that different mixture components
represent different \emph{internal models}, that is, models of
scenarios where the rules are different. To study this, we repeated
the calculations outlined above, measuring the presence of such
scenarios, rather than stochastic events. As discussed above, we identify 3 
scenarios in
this game where the rules of how to generate the next frame are very
different from the normal situation (facing monsters and any
fireballs): 1) having a wall on the left, 2) having a wall on the
right and 3) getting hit by an exploding fireball. 


\begin{figure}[ht]
  \vskip 0.2in
  \begin{center}
    \centerline{\includegraphics[width=\columnwidth]{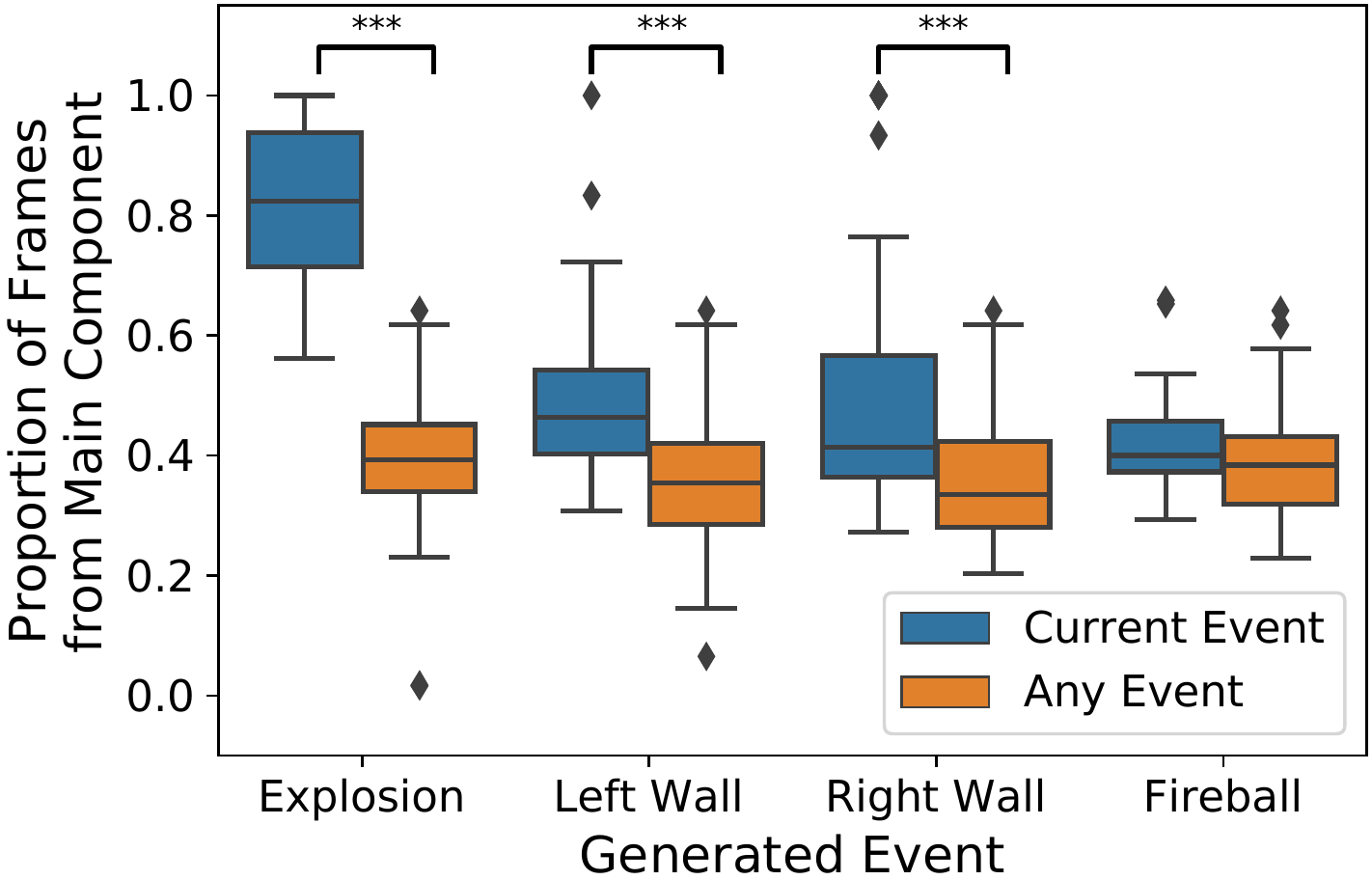}}
    \caption{The tendency for different scenarios to be produced by one 
      specific Gaussian component (blue) vs the tendency for that component 
      to be 
      responsible for events overall (orange).~$***$~indicates significant 
      differences with $p<0.001$.}
    \label{fig:world_proportions}
  \end{center}
  \vskip -0.2in
\end{figure}

The result of this calculation is shown in
Figure~\ref{fig:world_proportions}. There are statistically
significant differences ($p<0.001$) between the main component's
tendency to generate the specific situations and their tendency to
generate frames overall, for explosions and walls on either side. We
also show that the same effect is not generally present for situations
containing fireballs. We hypothesize that this is because fireballs
are very common, and do not drastically change the way the world
changes from one frame to the next. There should therefore be less
need for modeling them in a separate mixture
component.




\subsection{Plotting component weights and dreams}

\begin{figure}
  \centering
  \begin{subfigure}[b]{0.49\textwidth}
    \includegraphics[width=\textwidth]{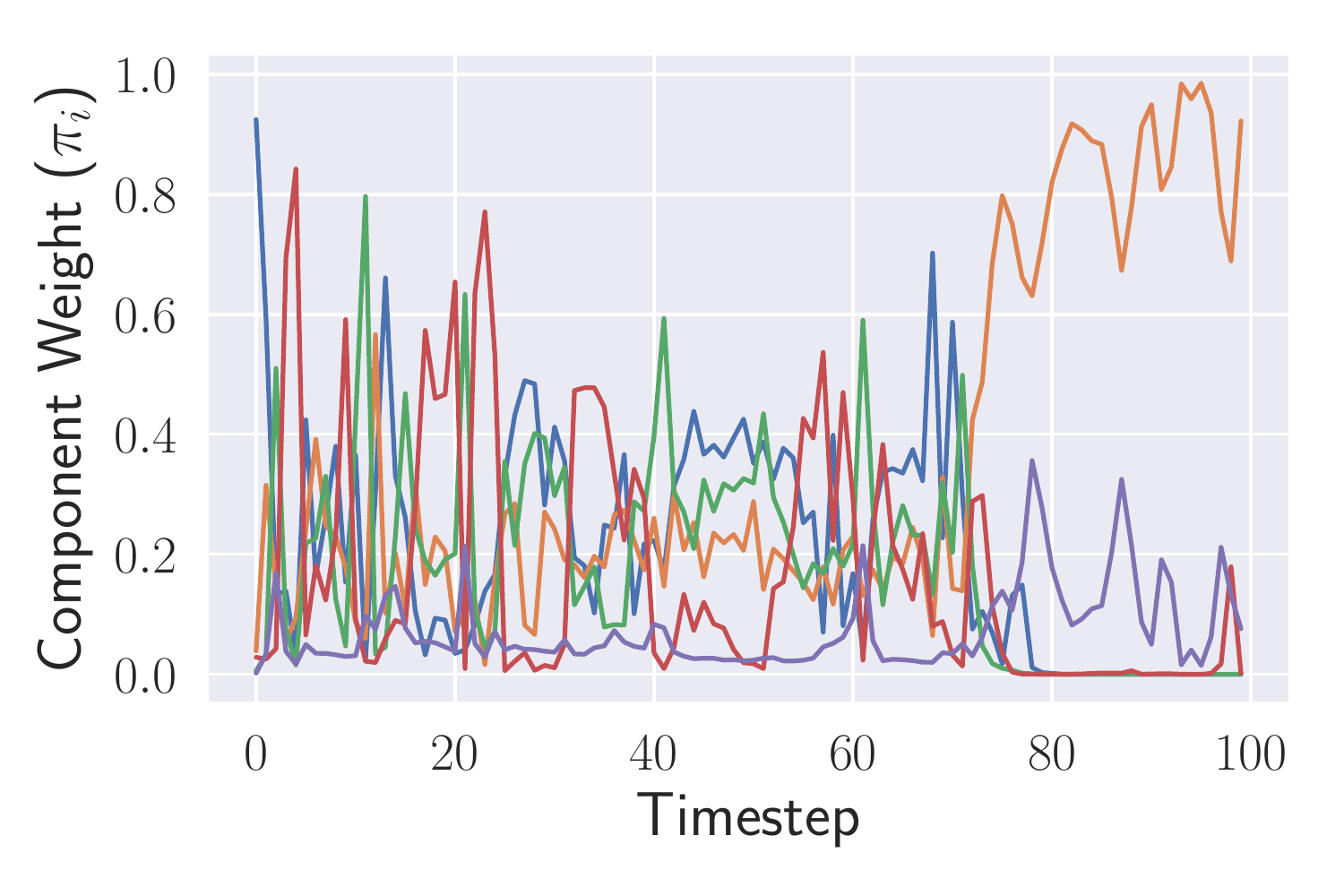}
    \caption{The weights of the 5 different mixture components ($\pi_i$ for $i 
    \in 
    \{1,2,3,4,5\}$) during a 100-timestep 
    dream.}
    \label{fig:mixture_weights}
  \end{subfigure}
  \hfill
  \begin{subfigure}[b]{0.49\textwidth}
    \includegraphics[width=\textwidth, trim={1cm 0 0cm      
    0},clip]{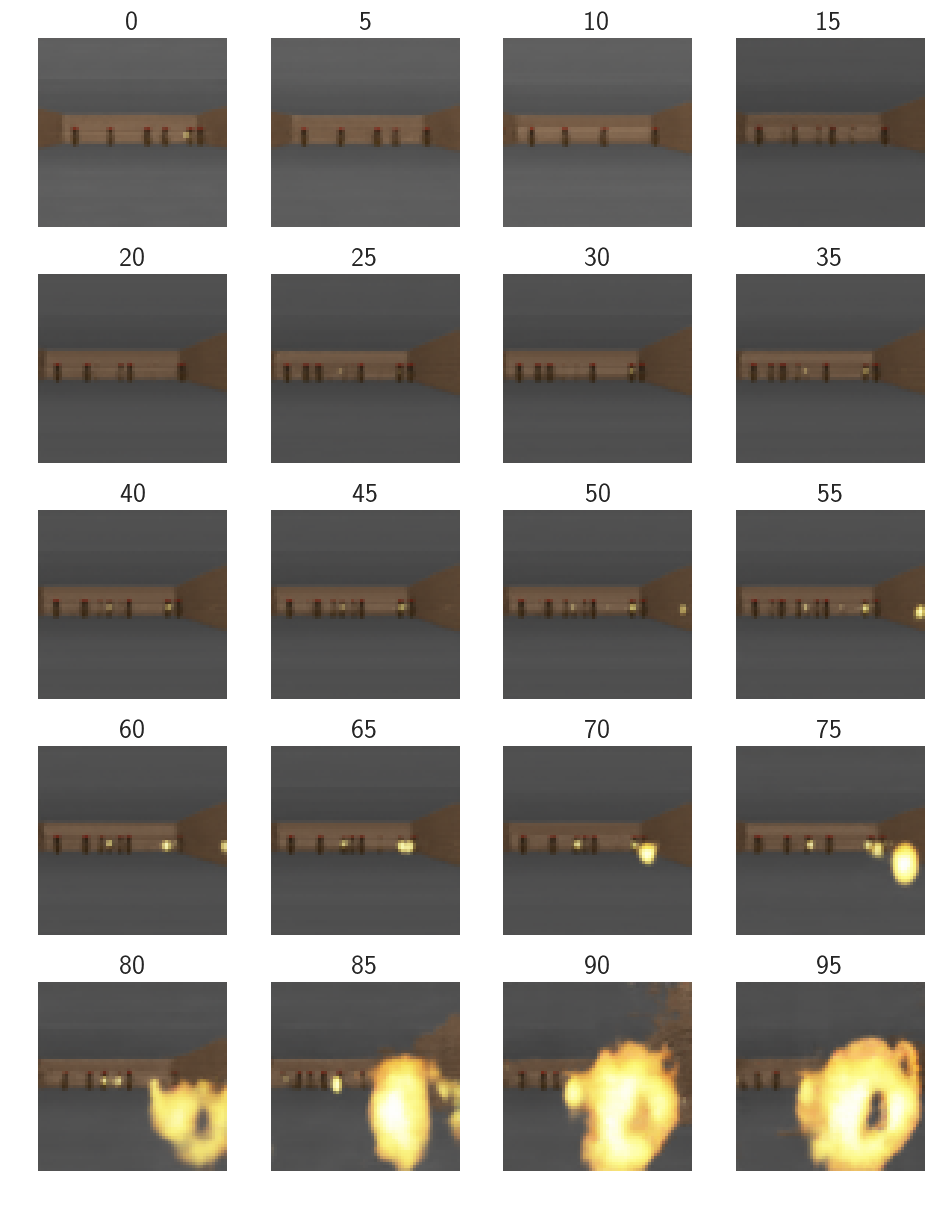}
    \caption{The resulting dream generated by sampling according to the mixture 
    weights in (a). Numbers indicate the current timestep.}
    \label{fig:rollout}
  \end{subfigure}
  \caption{The weights over time of each of the 5 mixture components output by 
  the MD-RNN and the 
  corresponding dreams produced by sampling according to those weights. Until 
  timestep 75, several components are similarly weighted, and responsible for 
  making predictions together. After timestep 75, one component dominates, and 
  takes over in generating predictions. The resulting prediction is an 
  exploding fireball.}
  \label{fig:weights_and_rollout}
\end{figure}

To further illuminate the role of different mixture components, we
let a single trained MD-RNN dream up 100-timestep predictions, while
plotting the weights of all components, to see which one is currently
most responsible for making predictions. An example of such a
plot is shown in Figure~\ref{fig:weights_and_rollout}. Notice
one specific mixture component dominates from around timestep 75, the
same time that an explosion is present in the frame. In other
repetitions of the experiment, we found a different component tends to
dominate when the agent is near a wall. In ``normal'' situations (no
nearby walls or explosions), we tend to see different components being
active together, without any clear dominance. This supports our second
hypothesis, that different components specialize to model scenarios
where the rules for generating future frames are different.


\subsection{Committing to one mixture}

As a final test of the role of different mixture components in
making predictions, we generated dreams while \emph{committing} to a single
component during an entire dream. We conditioned the MD-RNN with a
random start image, and made a dream predicting 1000 steps into the
future, sampling only from the \emph{first mixture component}. We then repeated
this for each of the five
mixture components in the MD-RNN. Since different trained models may not
represent the same events in the exact same mixture components, we
base this analysis on ten 1000-timestep dreams for each component, from
a single trained model.

\begin{figure}[ht]
  \vskip 0.2in
  \begin{center}
    \centerline{\includegraphics[width=\columnwidth]{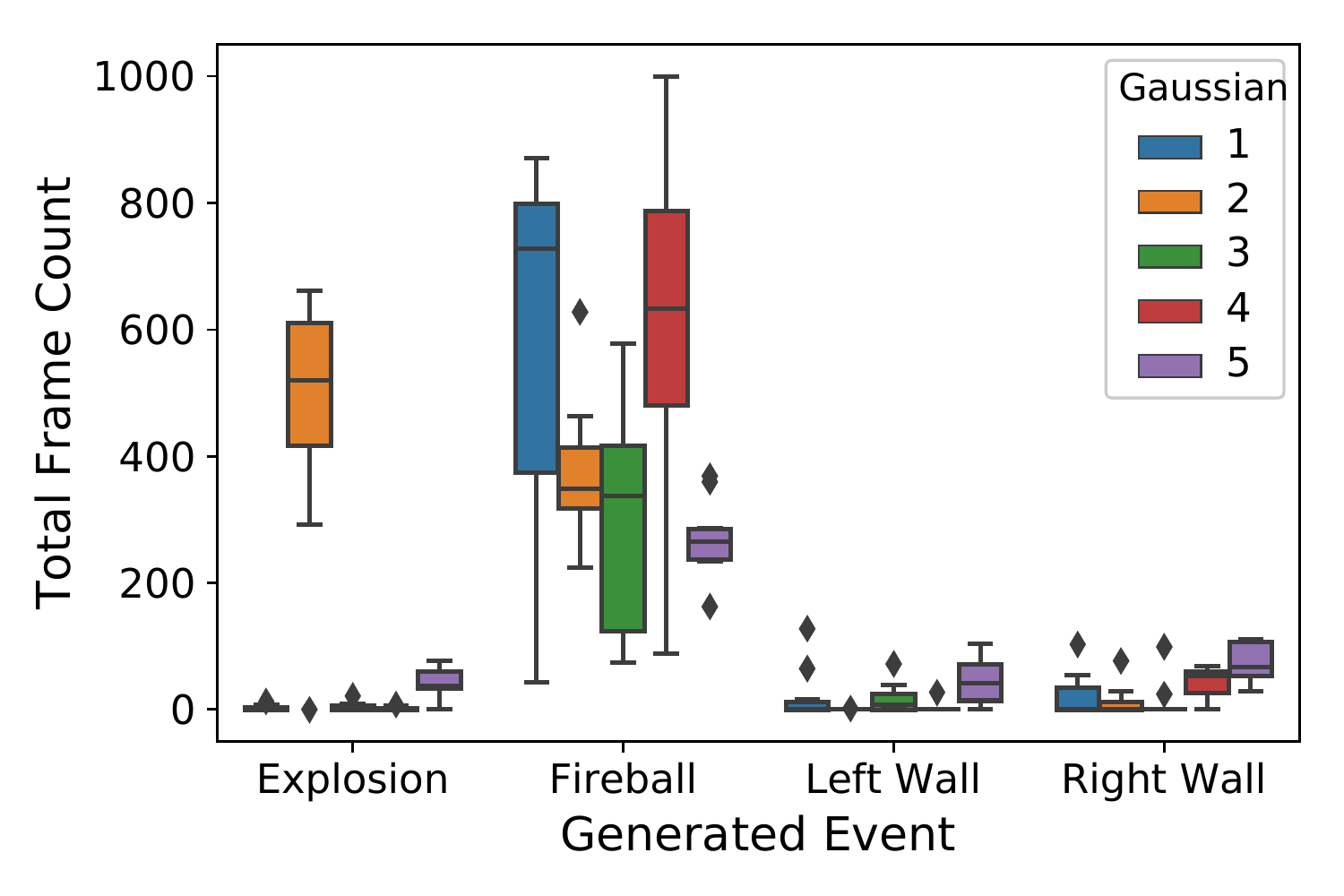}}
    \caption{The events generated in dreams, when committing to a single 
    Gaussian component during the entire prediction sequence.}
    \label{fig:committed_dreams}
  \end{center}
  \vskip -0.2in
\end{figure}

The results are shown in Figure~\ref{fig:committed_dreams}. There is a
clear tendency for this model to generate explosions with the second
mixture component, and walls (both right and left) with the fifth component.
Visualizing the conditional dreams, we observe something interesting: The 
components that do not produce explosions result in dreams where fireballs 
approach the
player, but stop and hover mid-air, or even reverse and return to the
monsters. Presumably, these components have never learned to model
explosions, and can therefore not produce them when being responsible
for generating dreams alone.

\section{Discussion}

The results give support for both our initial hypotheses: The different 
Gaussian components in the MD-RNN specialize to model both different stochastic 
events (Figure~\ref{fig:event_proportions}) and different internal models 
(Figure~\ref{fig:world_proportions}). For the stochastic events, we saw a 
strong tendency for fireball appearances to be generated more frequently by a 
specific mixture component, but the same was not true for monster appearances. 
Detecting monsters in the generated predictions is more difficult than the 
other elements we detect (see Appendix section 2), and we cannot rule out that 
a relationship exists here that we could have measured if monster appearances 
were less ambiguous.

For the second hypothesis, we observed very clear evidence that situations 
where the rules governing predictions are different were produced by separate 
Gaussian components. This was seen clearly both when measuring which components 
were most active in the different situations 
(Figures~\ref{fig:world_proportions} and~\ref{fig:weights_and_rollout}), and 
when sampling from a single 
component during an entire prediction sequence 
(Figure~\ref{fig:committed_dreams}).


\section{Conclusion}

Through automatic classification of predicted frames, we have shed
light on the way mixture density RNNs predict the future. We started
out with two hypotheses for the role of the different components in
the Gaussian mixture models, and found some evidence in support of
both. First, we found evidence that different components
 produce different stochastic events more frequently,
supporting the hypothesis that different components of the mixture
models represent different potential directions for the predicted
future. This is a valuable property for systems modeling creative
predictions (such as in generation of artistic text, music and
images), since it allows them to strike a balance between modeling an
observed phenomenon and improvising by choosing between several
possible predicted futures.

We found even more solid evidence for our second hypothesis: There is
a very strong tendency for different components of the mixture model
to be responsible for producing events that are governed by different
``rules'', that is, events that require different internal models.
Building machine learning systems that can represent different
internal models is a long-standing challenge, since learning of very
different skills tends to cause interference or
forgetting~\cite{Ellefsen2015}. One way this challenge has been
handled in the past, is by building modular systems, where different
modules \emph{compete} to represent different internal
models~\cite{Demiris2006, Haruno2001}. Our results suggest that
mixture density RNNs self-organize to separate different internal
models into different components.

This ability of MD-RNNs opens up a further hypothesis: Since these
networks can automatically self-organize multiple internal models,
they should be well equipped to model different scenarios with a low
degree of interference. In future studies, we plan to examine this
further by training MD-RNNs on multiple different
environments, studying the effect of the number of components in the
MDN on the observed interference.

\section{Acknowledgments}

This work is partially supported by The Research Council of Norway as a part of 
the Engineering Predictability with Embodied Cognition (EPEC) project, under 
grant agreement 240862.

\pagebreak

\bibliographystyle{icml2019}
\bibliography{references.bib}

\end{document}